\documentclass{article}

\usepackage[preprint]{neurips_2024}

\usepackage{enumitem}

\usepackage[utf8]{inputenc} 
\usepackage[T1]{fontenc}    
\usepackage{hyperref}       
\usepackage{url}            
\usepackage{booktabs}       
\usepackage{amsfonts}       
\usepackage{nicefrac}       
\usepackage{microtype}      
\usepackage{xcolor}         
\usepackage{amsmath} 
\usepackage{tabularx}
\usepackage{graphicx}
\usepackage[most]{tcolorbox} 

\usepackage{natbib}
\setcitestyle{numbers,square}

\title{TIE: Revolutionizing Text-based Image Editing for Complex-Prompt Following and High-Fidelity Editing}

\author{%
  Xinyu Zhang\thanks{Equal contribution.}\\
Meituan Inc. \\
  \texttt{xinyu-zhang@outlook.com} \\
  \And
  Mengxue Kang\footnotemark[1] \\
  Meituan Inc. \\
  \texttt{kangmengxue@hotmail.com} \\
  \AND
  Fei Wei \\
  Meituan Inc. \\
  \texttt{fwei\_mail@163.com} \\
  \And
  Shuang Xu \\
  Meituan Inc. \\
  \texttt{sxu1997@126.com} \\
  \And
  Yuhe Liu \\
  Beihang University \\
  \texttt{liuyuhe@buaa.edu.cn} \\
  \And
  Lin Ma \thanks{Corresponding Author.} \\
  Meituan Inc. \\
  \texttt{forest.linma@gmail.com} \\
}

\begin{document}

\maketitle

\begin{abstract}
As the field of image generation rapidly advances, traditional diffusion models and those integrated with multimodal large language models (LLMs) still encounter limitations in interpreting complex prompts and preserving image consistency pre and post-editing. To tackle these challenges, we present an innovative image editing framework that employs the robust Chain-of-Thought (CoT) reasoning and localizing capabilities of multimodal LLMs to aid diffusion models in generating more refined images. We first meticulously design a CoT process comprising instruction decomposition, region localization, and detailed description. Subsequently, we fine-tune the LISA model, a lightweight multimodal LLM, using the CoT process of Multimodal LLMs and the mask of the edited image. By providing the diffusion models with knowledge of the generated prompt and image mask, our models generate images with a superior understanding of instructions. Through extensive experiments, our model has demonstrated superior performance in image generation, surpassing existing state-of-the-art models. Notably, our model exhibits an enhanced ability to understand complex prompts and generate corresponding images, while maintaining high fidelity and consistency in images before and after generation.
\end{abstract}

\vspace{0.25cm}
\section{Introduction}
\vspace{0.25cm}
The emergence of generative models has marked a significant milestone in the field of artificial intelligence, particularly in the domain of computer vision \cite{ho2020denoising,dhariwal2021diffusion,rombach2022high,ho2022classifier,zhang2023adding,liu2023llava}. A category of these models, known as text-guided image models, have demonstrated remarkable capabilities in creating images from varied text prompts \cite{ruiz2023dreambooth,gal2022image, nichol2021glide,brooks2023instructpix2pix,hertz2022prompt}. These models are adept not only at generating images that align with the text, but also at positioning unrelated objects in visually appealing ways. This advancement underscores an era where the barriers between textual descriptions and visual representations are increasingly blurred, allowing machines to create with a degree of detail that was previously exclusive to human imagination.

However, the expectations and demands for such models extend far beyond the execution of simple commands. Users are seeking the ability to convey complex, nuanced instructions that require a deeper level of comprehension and precision. For instance, a prompt such as "Turn the hair of the person on the left red, and transform the dress of the person on the right into a white sundress" poses a significant challenge for pure text-conditional image generation models. While these models may recognize individual attributes such as colors and objects, their capacity to associate these attributes with the correct objects is limited. Furthermore, their proficiency in capturing all aspects of the corresponding text prompt is often inadequate. As a result, such complex instructions remain out of reach for traditional text-conditional image generation models, creating a disconnect between user intentions and model performance.

\begin{figure}[t]
\hspace{-0.5cm}
\centering
\includegraphics[scale=0.45]{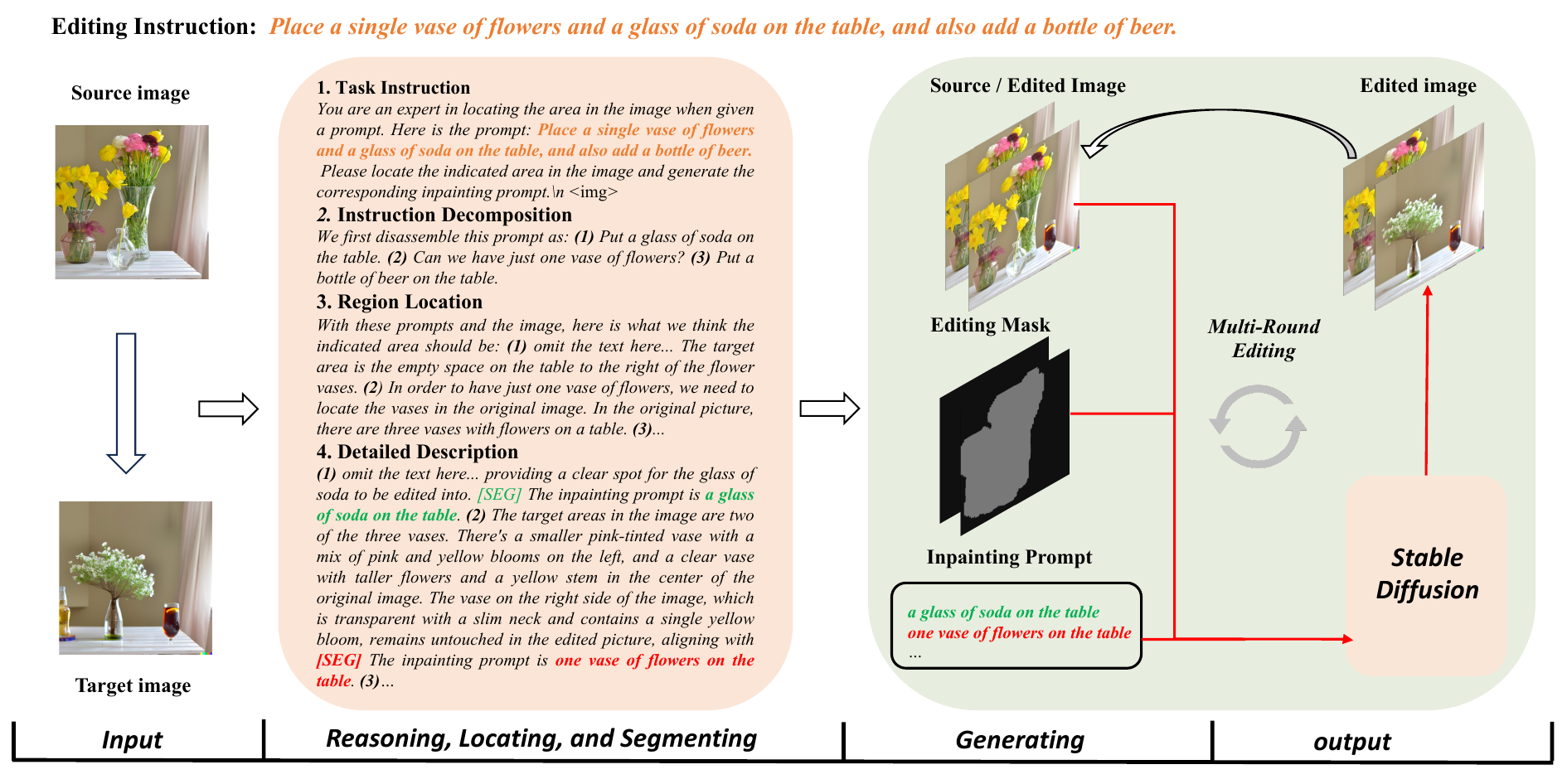}
\caption{Illustration of our proposed framework as a Chain-of-Thought (CoT) process.}
\label{fig:pipeline}
\end{figure}

Conversely, generative models powered by multimodal large language models (MLLM) have made considerable advancements. By harnessing the strong language comprehension capabilities of large language models (LLM), these systems offer a nuanced interpretation of given instructions. Models such as Emu2 \cite{sun2023generative2} and Seed-X \cite{ge2024seed} augment text-conditional image generation models by incorporating feature embeddings, while SEED-LlaMA \cite{ge2023making} integrate features extracted from codebooks. However, there is a misalignment between the features and the latent space required for stable diffusion, resulting in suboptimal generated images that display noticeable deviations from the intended outcome. Another category of models, such as Mini-Gemini \cite{li2024mini}, utilizes MLLM to generate text, which is then input into a text-conditional image model capable of effectively comprehending the input. However, the images produced by this process are typically coarse and exhibit significant discrepancies from the original images.

These observations highlight two prevalent challenges across both traditional text-conditional image models and MLLM-based image generation: (1) A lack of robust understanding of complex instructions, particularly when it comes to disambiguating attributes within a prompt or fully realizing specific parts of an instruction. (2) An inability to generate images with high fidelity that retain the essence of the original picture, especially when MLLM-generated embeddings or text are provided to a text-conditional image model, resulting in outputs that diverge considerably from the source material. Addressing these challenges is crucial for advancing the field and meeting the growing demands for more sophisticated and accurate image generation from textual descriptions.

To address the aforementioned limitations in image synthesis from complex textual prompts, we introduce a novel generative framework that bridges multimodal language and image learning models (MLLM) with image generation. This innovative approach leverages the impressive reasoning capabilities and instance-level segmentation prowess of multimodal LLMs to enhance the controllability of diffusion models. Our model operates in three distinct phases:

At the core of our framework lies the prompt reasoning process, a critical phase where a fine-tuned LISA \cite{lai2023lisa} model, informed by chain-of-thought (CoT) data generated by GPT-4V \cite{achiam2023gpt}, is employed. This process is pivotal as it equips the model with the ability to dissect and comprehend intricate prompts, thereby setting the stage for precise image manipulation. The LISA model, with its finetuned parameters, adeptly navigates through the nuanced intricacies of the prompts, ensuring that the subsequent steps are based on a solid understanding of the task at hand. The second pillar of our framework is the LISA localizing process, a sophisticated mechanism that utilizes advanced techniques (to be elaborated upon) to generate masks for objects specified within the text. This process is not merely about identifying the objects in question but also about understanding the directional cues embedded within the prompts, which dictate the necessary alterations. The resulting masks serve as a blueprint, guiding the model to focus on the regions of interest, thereby streamlining the generation process and ensuring that the modifications are confined to the intended areas. Finally, the inpainting process \cite{kandinsky2.2} brings together the reasoned prompts and the meticulously generated masks to breathe life into the new images. This phase is where the magic happens, as our model synthesizes images that not only adhere to the complex prompts but also mirror the fidelity of the original images. By harmoniously blending the input prompt and the corresponding mask, the model is able to execute precise edits, thus producing high-fidelity images that are a true reflection of the user's intent while preserving the essence of the source material.

In summary, below are the contributions of this paper:
\begin{itemize}[leftmargin=0.6cm]
\item We innovatively propose a comprehensive image generation model that follows complex instructions and produces high-fidelity images, maintaining the integrity of the original image.
\item We firstly use small language model to perform the CoT process of reasoning and localizing, ensuring the model's robust reasoning capabilities while significantly reducing operational costs.
\item We introduce a novel methodology that utilizes text generated by MLLM and image masks generated by LISA to inform the diffusion model, resulting in superior synthesis quality and guaranteeing high-fidelity image generation.
\item We provide a new pipeline dataset, which includes the source images, mask images, target edited images, complex instructions and the CoT process, offering a holistic resource for advancing image synthesis research.
\end{itemize}

\section{Related works}
\par \noindent \textbf{Text-guided Image Generation.} Text-to-Image (T2I) generation has gained popularity and become one of the most hot topics currently. Early T2I methods are based on GANs \cite{xu2018attngan,zhu2019dm}, which trained a generator and a discriminator in the adversarial process. Another stream of T2I methods follow an auto-regressive pipeline \cite{ramesh2021zero,ding2021cogview,yu2022scaling}, mainly predicting image tokens sequentially by employing Transformers. More recently, Diffusion models \cite{nichol2021glide,saharia2022photorealistic,SD, ramesh2022hierarchical} have made great progress compared to GANs and auto-regressive methods due to their ability to generate highly realistic and more diverse images.
GLIDE \cite{nichol2021glide} is the first T2I framework based on the Diffusion Model, which replaced the original class label with text. Imagen \cite{saharia2022photorealistic} follows GLIDE and adopts a pretrained and frozen large language model as the text encoder. LDM \cite{rombach2022high} compresses images into low-dimensional latent space representations, effectively reducing computational complexity. Stable Diffusion \cite{SD} is a milestone work which scaled up based on LDM \cite{rombach2022high}. DALLE-2 \cite{ramesh2022hierarchical} takes CLIP \cite{radford2021learning} as the text encoder and generates images from the CLIP latent space. The progress in diffusion models has also stimulated their applications in text-to-image editing. For instance, Prompt-to-Prompt \cite{hertz2022prompt} facilitates image editing by altering words in the original prompts and incorporating cross-attention maps during the diffusion process. DiffEdit \cite{couairon2022diffedit} emphasizes areas of an input image that should be modified based on a text query to assist editing. InstructPix2Pix \cite{brooks2023instructpix2pix} provides a large-scale editing dataset composed of instruction-based samples. HQ-Edit \cite{hui2024hq} further finetunes InstructPix2Pix, enhancing the performance of cutting-edge text-to-image editing.


\par \noindent \textbf{Multi-modal LLM based Understanding and Generation.}
In light of the rapid advancement and impressive performance of recent multimodal large language models, numerous studies have begun to explore multimodal large language model based understanding and generation models. These methods can be broadly categorized into two types: embedding-based and text-based approaches. Embedding-based approaches utilize visual embeddings extracted from images as autoregressive targets for multimodal models. \cite{yu2023scaling, jin2023unified, ge2023making, team2023gemini, lu2023unified} employ a visual tokenizer to encode images into discrete visual embeddings. For instance, \cite{lu2023unified} utilizes a densely pre-trained VQ-GAN model to encode images into token embeddings with a codebook. \cite{ge2023making} introduces SEED, a VQ-based image tokenizer, to enhance the capability of LLMs to simultaneously handle understanding and generation tasks. Furthermore, \cite{sun2023generative, dong2023dreamllm, zhu2023vl, sun2023generative2, ge2024seed} extract visual features from images through a visual encoder to obtain continuous visual embeddings. DreamLLM expands a special \texttt{<dream>} token to predict where to generate images. SEED-X constrains the reconstruction of images to be semantically aligned with the original images, reducing model distortion and enhancing the realism of generated image details. Text-based approaches generate corresponding text prompt instructions via large-scale models, which are then input into generative models. For example, \cite{li2024mini} leverages VLM guidance for image generation by providing the generated text from LLMs.

\section{Methods}


In this section, we formulate our proposed framework as a Chain-of-Thought process, which is distinguished by complex prompt decomposition, reasoning and refinement for high-fidelity image editing. First, we utilize a multi-modal LLM to interpret a complex text-based editing instruction with the input image, and decompose it into several basic editing instructions. Second, we use the model to identify the target region in the input image referred to by these instructions, and create detailed descriptions of the corresponding area, aiding in generating precise mask output. Meanwhile, a semantically-aligned inpainting prompt is generated by imaging the content of the edited image in the mask area. Finally, we input the input image, generated mask and text prompt into an inpainting model to generate a high-fidelity image.


\subsection{Chain-of-Thought Pipeline}
Here, we introduce a Chain-of-Thought (CoT) process to analyze the user's editing instructions. This analysis unfolds through a tripartite process, \textit{i.e.,} \textbf{\textit{instruction decomposition}}, \textbf{\textit{region localization}}, and \textbf{\textit{detailed description}}.

\begin{table}[b]
\centering
\begin{tcolorbox}[colframe=gray, 
                  colback=white!94!black, 
                  arc=4mm, 
                  boxsep=5mm, 
                  boxrule=1pt] 
\textit{ADD:} Insertion of an object into the image. It can identify the appropriate context within the image and seamlessly integrate a new object, maintaining the natural aesthetics of the scene. \\

\textit{REMOVE:} Erasure of an object from the image. It can accurately recognize and remove the specified object, subsequently filling in the resulting space with contextually relevant content that matches the surrounding area. \\

\textit{CHANGE:} \\
 \textit{Object}: Replacing one object with another, ensuring that the new object harmonizes with the existing environment in the image. \\
 \textit{Attribute}: Changing an attribute such as color or texture. It can apply these changes to the specified object without affecting the rest of the image. \\
 \textit{Background}: Modifying the background of the image while preserving the foreground. 
\end{tcolorbox}
\caption{Editing Operations supported in our method.}
\label{tab:instructions}
\end{table}

\textbf{\textit{Instruction Decomposition.}} In the initial phase, the multi-modal LLM decomposes the complex editing instruction into multiple editing instructions, and summarize these instructions as several simpler sub-prompts. Note that each sub-prompt is confined to a single operation associated to the corresponding target. For example, a compound instruction like 
 \textit{REMOVE A and ADD B} is dissected into two simpler prompts: \textit{REMOVE A}, \textit{ADD B}. This decomposition allows for a more granular and manageable approach to image editing.  
We summarize a list of supported image editing instructions in Tab. \ref{tab:instructions}.


\textbf{\textit{Region Localization.}} In this phase, the model embarks on an analytical reasoning process for each simplified sub-prompt. It can identify the specific region of the image that corresponds to each operation. Specifically, \textit{REMOVE} operation targets a particular foreground object, \textit{CHANGE} operation also focuses on a specific foreground object, whereas \textit{ADD} operation is concerned with a particular background object. Taking the instruction \textit{{ADD a dog on the sofa}} as an example, the target region for modification would be the sofa, the background area upon which the new object, a dog, is to be superimposed. Subsequent image generation will be performed in this background area to seamlessly integrate the new object into the scene.

\textbf{\textit{Detailed Description.}} This phase requires the model to furnish an exhaustive characterization of the area in the editing instruction. This includes articulating attributes such as the area's relative position within the image, its color, shape, and any other salient features that define its uniqueness. By delivering such a detailed description, the model aids in the precise identification of the intended area for editing. In addition, the model, taking into account the mask area and surrounding image content, imagine the content to be filled in the mask area of the post-editing image, and consequently generates a semantically aligned prompt.

\subsection{Chain-of-Thought Fine-tuing}

\subsubsection{Data Preparation}

\begin{figure}[h]
\centering
\includegraphics[width=1.0\textwidth]{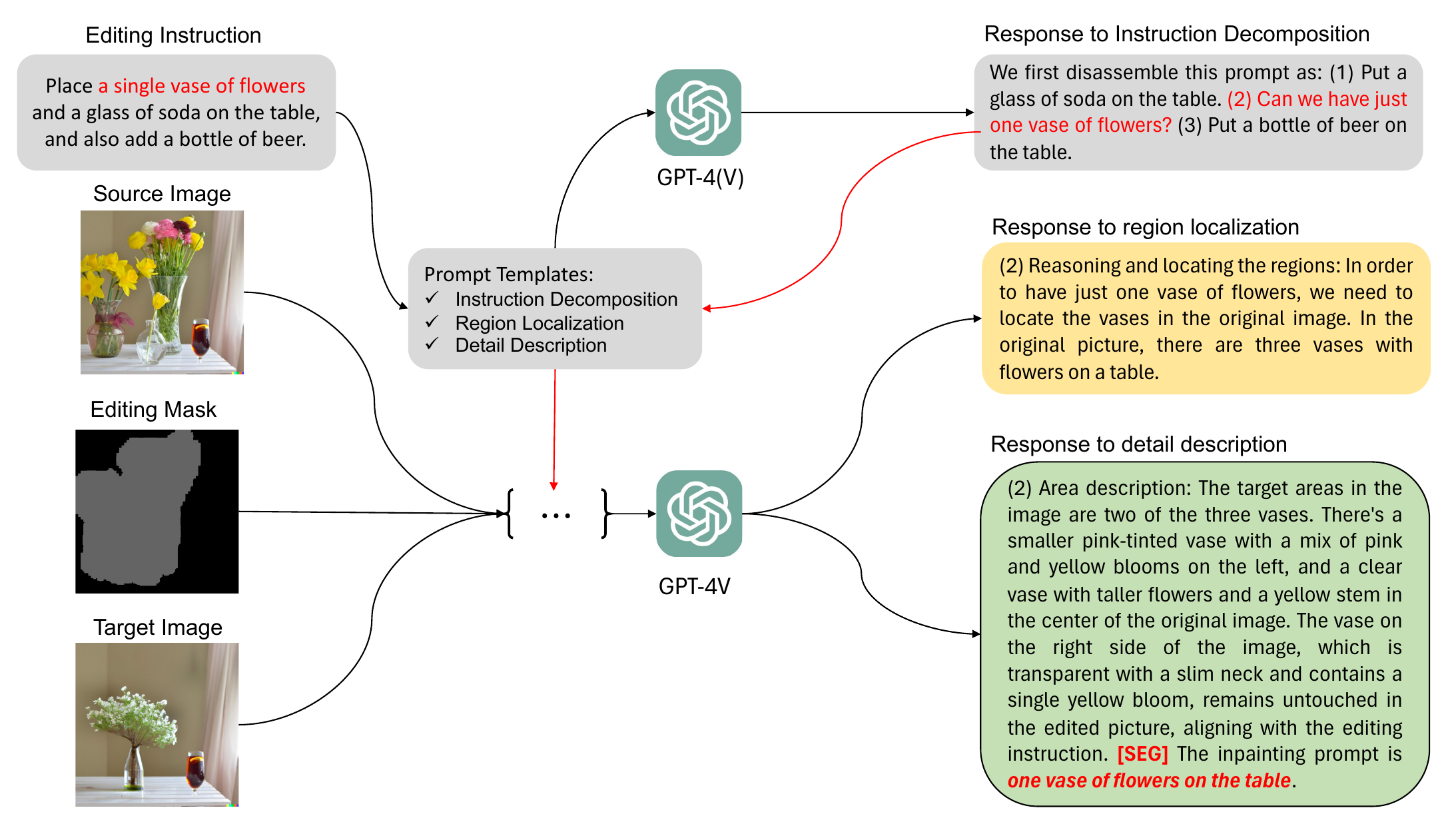}
\caption{Illustration of data preparation for CoT fine-tuning.}
\label{fig:data_preparation}
\end{figure}

In this work, we underscore the importance of data curation, which we initiate by leveraging the GPT-4v model to generate the Chain-of-Thought (CoT) process. Given original image $\mathcal{X}_{i}$, maksed image $\mathcal{M}_{i}$, target image $\mathcal{T}_{i}$ and editing instruction $\mathcal{I}_{i}$, we utilize GPT-4v to generate CoT Responses $\mathcal{R}_{i}$ corresponding to the \textit{instruction decomposition}, \textit{region localization}, and \textit{detailed description} processes. We detail prompt templates of generating these processes with GPT-4V in Tab.\ref{tab:data_generation}. For instance, we sent the original image, masked image, target image, and several prompts into GPT-4V, and resulting in several responses as shown in Fig.  \ref{fig:data_preparation}. Then, we encapsulate $\mathcal{X}_{i}$, $\mathcal{I}_{i}$, $\mathcal{M}_{i}$, $\mathcal{T}_{i}$, and $\mathcal{R}_{i}$ into a sample, which is represented as $\mathcal{S}_{i}$:
\begin{equation}
\mathcal{S}_{i} = \{\mathcal{X}_{i}, \mathcal{I}_{i}, \mathcal{M}_{i},  \mathcal{T}_{i}, \mathcal{R}_{i}\}
\end{equation}
This sample is subsequently used for fine-tuning the LISA model, enabling it to simultaneously generate a precise mask area based on detailed area descriptions, and a better inpainting prompt that well aligns with the editing instructions.


\subsubsection{Model Fine-tuning}

In this work, we formulates the image editing task as an image inpainting task that takes precise target masks and well-aligned text prompts. Considering that LISA inherits the language generation capabilities of multimodal LLMs and also shows the excellent ability to produce segmentation masks, we use it to perform the CoT process. However, limited to the relatively small capacity of language model, LISA cannot perform as well as GPT-4V in the area of in-context learning. Therefore, we leverage the reasoning samples generated by GPT-4V using CoT prompting and subsequently fine-tune LISA using these samples, which is known as Fine-tune-CoT \cite{ho2022large}. This strategy retains the flexibility of prompt-based CoT methods while mitigating their dependence on prohibitively large models. Inspired by segmentation in dialogue (SiD) proposed in LISA++ \cite{yang2023improved}, we reformat these samples in the style of segmentation in CoT as following:

\begin{itemize}[leftmargin=0.6cm]
    \item \textbf{USER}: <img> \textbackslash n You are an expert in locating the area in the image when given a prompt. Here is the prompt: \textit{Place a single vase of flowers and a glass of soda on the table, and also add a bottle of beer}. please locate the indicated area in the image and generate the corresponding inpainting prompt.
    \item \textbf{ASSISTANT}: We first disassemble this prompt as: (1) Put a glass of soda on the table. (2) Can we have just one vase of flowers? (3) Put a bottle of beer on the table.  \textbackslash n With these prompts and the image, here is what we think the indicated area should be: (1) - Reasoning and locating the regions: \textit{\{omit the text here...\}} providing a clear spot for the glass of soda to be edited into. \textcolor[RGB]{0,200,0}{\textbf{[SEG]}} The inpainting prompt is \textbf{\textit{\textcolor[RGB]{0,200,0}{a glass of soda on the table}}}. (2) - Reasoning and locating the regions:\textbackslash n In order to have just one vase of flowers, we need to locate the vases in the original image. In the original picture, there are three vases with flowers on a table. \textbackslash n - Area description:\textbackslash n The target areas in the image are two of the three vases. \textit{\{omit the text here...\}} aligning with the editing instruction. \textcolor[RGB]{255,0,0}{\textbf{[SEG]}} The inpainting prompt is \textbf{\textit{\textcolor[RGB]{255,0,0}{one vase of flowers on the table}}}. (3) - Reasoning and locating the regions: \textit{omit the text here...}
\end{itemize}

Following the setting in LISA, we expand the LLM vocabulary with a new token, i.e., [SEG], which responses to the request for the segmentation, and feed its embedding into the decoder~\cite{kirillov2023segment} to produce the final segmentation mask as $\hat{\mathcal{M}}$. Meanwhile, we also extract the corresponding inpainting prompt $\hat{\mathcal{P}}$ from the model's output. Then, we sent the predicted mask $\hat{\mathcal{M}}$ and prompt $\hat{\mathcal{P}}$ for the relevant editing area into the powerful inpainting model Kandinsky-2.2-decoder-inpaint \cite{kandinsky2.2}. This approach ensures a comprehensive and accurate representation of the user's initial editing instruction, thereby enhancing the overall quality and relevance of the generated images.

\section{Experiments}
\subsection{Dataset}
MagicBrush \cite{NEURIPS2023_64008fa3} a large-scale and manually annotated dataset for instruction guided image editing. We extracted 1376 samples from the MagicBrush dataset to construct our new pipeline dataset. For the construction of the dataset, we first summarized multiple editing instructions into one complex instruction, and then used the CoT process proposed in the Section 3.2.1 to supplement reasoning results. The elements of the final dataset include: original image, mask image, target edited image, complex instructions, intermediate reasoning of the CoT process.

\subsection{Implementation Details}
\textbf{Training} We follow the training configuration of the original LISA model \cite{lai2023lisa}, such as training epochs, learning rate, losses, loss weights and trainable parameters, \textit{etc.} Note that, we only fine-tune the LISA-13B model on the aforementioned 1376 samples that involve the CoT process including instruction decomposition, region localization, and detailed description. We use Kandinsky-2.2-decoder-inpaint to perform image editing without tuning any parameters of the model.

\textbf{Evaluation} 
We adopt several evaluation metrics following previous methods \cite{ruiz2023dreambooth, hui2024hq} to assess the image editing quality. There are mainly three aspects, including CLIP-I to compute image fidelity, CLIP-T and Alignment \cite{hui2024hq} to reflect the text-to-image alignment and complex instruction following ability, and the Coherence introduced in \cite{hui2024hq} to assess the overall aesthetic quality of the generated image (such as coherence towards lighting, shadow and style). For the aforementioned metrics, higher scores represent better performance.

%

\section{Results}

Our research methodology involved a comparative analysis of our model with the state-of-the-art models in the field of AI image editing. We evaluated our model against two types of generative models: pure diffusion models, including InstructPix2Pix \cite{brooks2023instructpix2pix} and Prompt2Prompt\cite{hertz2022prompt}, DiffEdit \cite{couairon2022diffedit} models, and multimodal large language model (MLLM) based generative models, including Emu2 \cite{sun2023generative2}, MM-Interleaved \cite{tian2024mm}, and Seed-X \cite{ge2024seed} models.

\begin{figure*}[t]
\vskip 0.2in
\begin{center}
\centerline{\includegraphics[width=0.97\columnwidth]{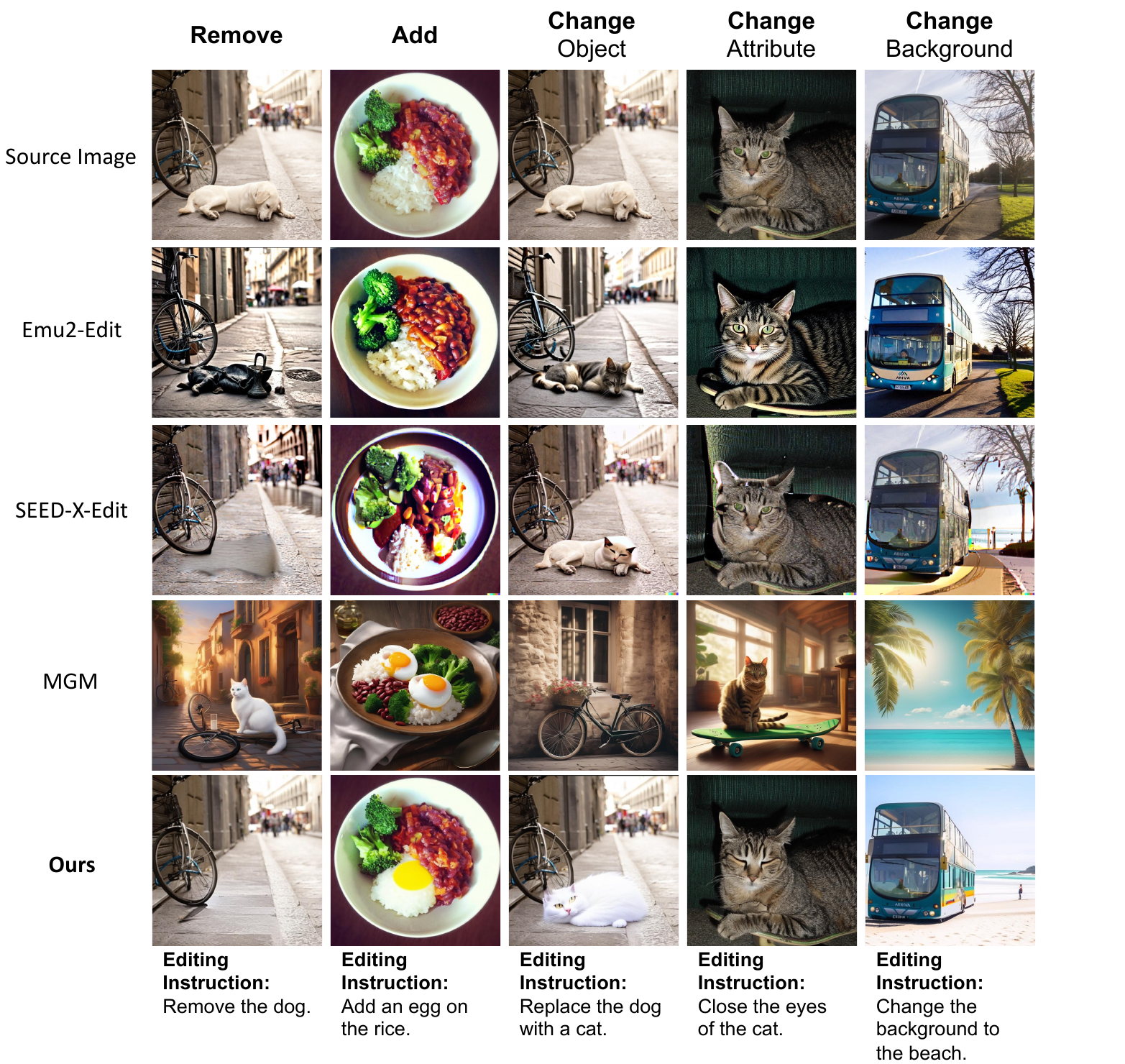}}
\caption{Visualization comparison with MLLM SOTA models.}
\label{fig:compare_with_sotas_mmllms}
\end{center}
\vskip -0.2in
\end{figure*}

\begin{table*}[t]
\centering
\begin{minipage}{0.45\textwidth}
\setlength{\tabcolsep}{2pt}
\centering
\begin{tabular}{lcccc}
\toprule
\multicolumn{1}{c}{} & \multicolumn{1}{c}{CLIP-T} & \multicolumn{1}{c}{CLIP-I} & \multicolumn{1}{c}{Ali.} & \multicolumn{1}{c}{Coh.} \\
\midrule
Emu2-Edit                     & 0.211                               & \textbf {0.708}                              & 43                                     & \textbf{85}                                      \\
SEED-X                        & 0.216                               & 0.679                               & 48                                     & 75                                      \\
MGM                          & 0.226                               & 0.424                               & 33                                    & 84                                      \\
Ours                          & \textbf{0.233}                              & 0.664                            & \textbf{57}                                     & 80     \\
\bottomrule
\end{tabular}
\caption{Comparing different MLLM models.}
\label{tab:mllm}
\end{minipage}
\hfill
\begin{minipage}{0.45\textwidth}
\setlength{\tabcolsep}{2pt}
\centering
\begin{tabular}{lcccc}
\toprule
\multicolumn{1}{c}{} & \multicolumn{1}{c}{CLIP-T} & \multicolumn{1}{c}{CLIP-I} & \multicolumn{1}{c}{Ali.} & \multicolumn{1}{c}{Coh.} \\
\midrule
StableDiffusion      & \textbf{0.278}                     & 0.368                      & 27                            & 23                             \\
DiffEdit             & 0.194                      & 0.639                      & 48                            & 74                             \\
HQ-Edit              & 0.252                      & 0.551                      & 45                            & \textbf{81}                            \\
Instruct-pix2pix     & 0.240                      & \textbf{0.671}                      & 52                            & 77                             \\
Ours                 & 0.233                   & 0.664                     & \textbf{57}                          & \textbf{80}                    \\
\bottomrule
\end{tabular}
\caption{Comparing different SD models.}
\label{tab:sd}
\end{minipage}
\end{table*}

\begin{figure*}[ht]
\vskip 0.2in
\begin{center}
\centerline{\includegraphics[width=0.95\columnwidth]{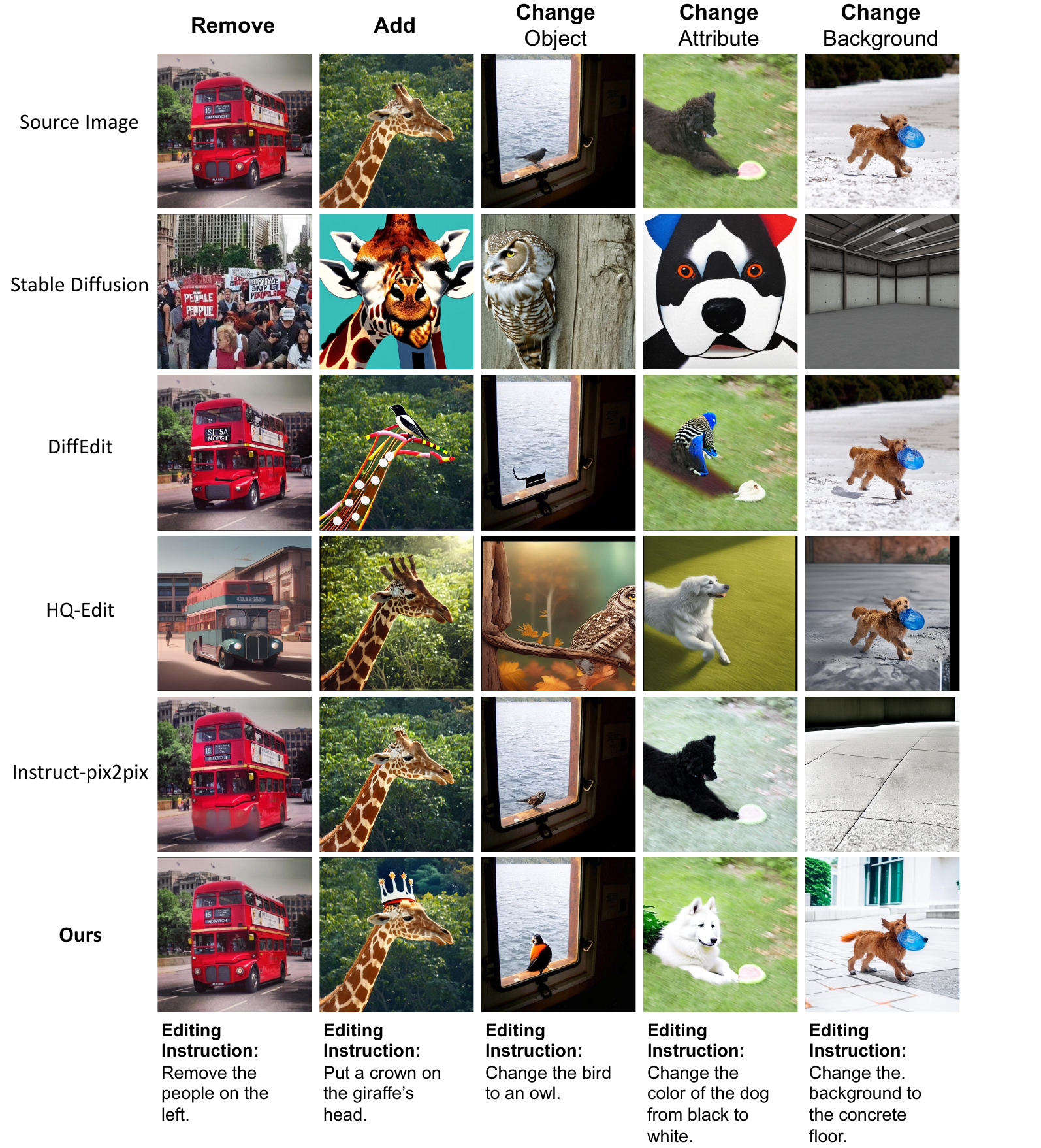}}
\caption{Visualization comparison with stable diffusion (SD) SOTA models.}
\label{fig:compare_with_sotas_sd}
\end{center}
\vskip -0.2in
\end{figure*}

\begin{figure*}[ht]
\vskip 0.2in
\begin{center}
\centerline{\includegraphics[width=0.9\columnwidth]{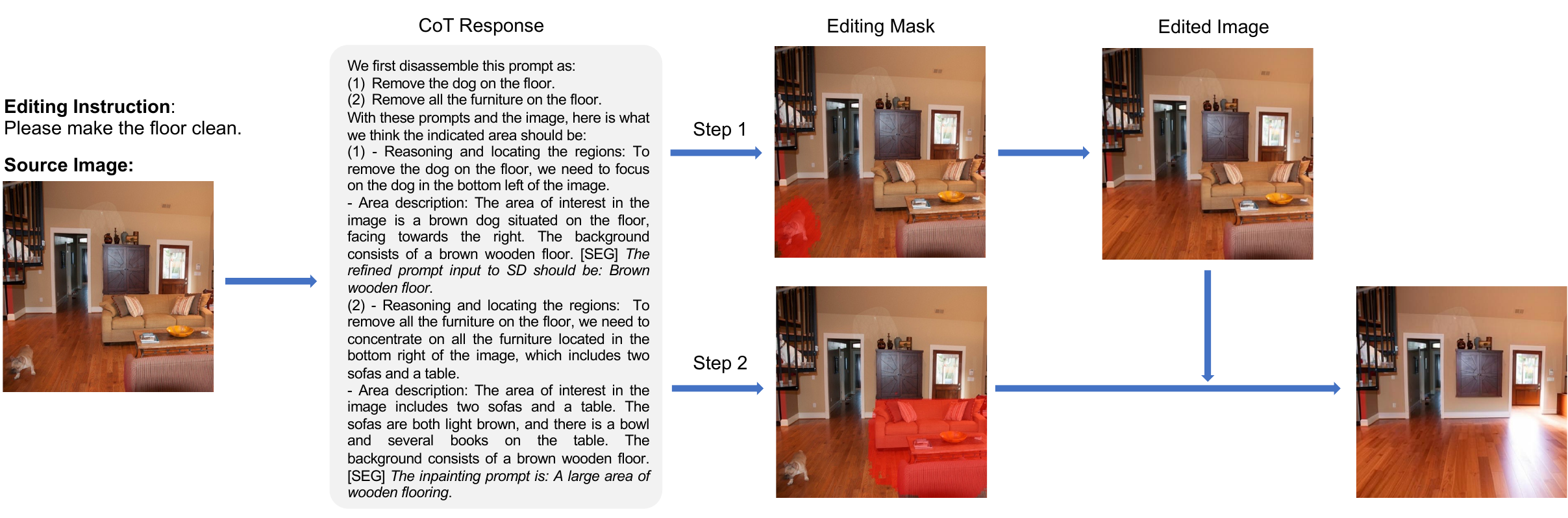}}
\caption{Image editing for complex-prompt.}
\label{fig:abl_complex}
\end{center}
\vskip -0.2in
\end{figure*}

The results, as depicted in Figure \ref{fig:compare_with_sotas_mmllms}, demonstrate that our model consistently produces more accurate editing results than the aforementioned models. The qualitative analysis of the image results indicates that other models struggle to fully follow the editing instructions. In contrast, our model accurately executes these operations, which we discuss in two primary aspects:

\textbf{Complex-prompt following.} Firstly, we discuss the ability of the models to follow complex instructions. Both the pure diffusion and MLLM models exhibit deficiencies in this regard. In the images generated by these models, some editing operations from the prompt are either lost or conflated. Our model effectively mitigates these issues, demonstrating a superior ability to follow instructions. From a quantitative perspective, we use the CLIP-T and alignment metrics in Tab.\ref{tab:mllm} and Tab.\ref{tab:sd}. Higher scores on these metrics indicate a closer adherence to the editing instructions in the generated images. Our model significantly outperforms others on these metrics, indicating a closer alignment with the editing prompts. 
We also provide an additional example to demonstrate our model's capability in text-guided image generation from complex prompts in Fig.\ref{fig:abl_complex}. Through the CoT process, our model can infer the necessary steps and handle a simple instruction at each step. After several iterations, it can generate an image that effectively follows the complex prompt.

\textbf{Fidelity of editing.} Secondly, we assess the fidelity of the models before and after editing. Both the pure diffusion and MLLM models generated images with significant discrepancies before and after editing. Ideally, the editing process should only alter a small region as specified in the prompt, with the rest of the image remaining unchanged. However, as illustrated in Fig.\ref{fig:compare_with_sotas_mmllms} and Fig.\ref{fig:compare_with_sotas_sd}, other models exhibit substantial differences between the pre-edited and post-edited images. In contrast, our model demonstrates meticulous alterations, modifying only the areas specified in the editing instructions. From a quantitative perspective, we use the CLIP-I and coherence metrics in Tab.\ref{tab:mllm} and Tab.\ref{tab:sd}. Higher scores on these metrics indicate smaller discrepancies between the pre-edited and post-edited images. Our model significantly outperforms others on these metrics, indicating that our image editing process minimizes discrepancies and adheres closely to the requirement of modifying only a specific region.



\section{Ablation Studies}

\textbf{Effectiveness of CoT process.} We compared the image generation outcomes with and without the implementation of CoT. 
The first two images were generated by LISA-13B \cite{lai2023lisa} model by directly inferring the corresponding mask from the user prompt and generating the image with the mask and original prompt, without any CoT process. In contrast, the latter two images were generated by our model, LISA-13B-sft, which undergoes a CoT process to generate a corresponding mask and a re-prompt for the area to be modified. The mask and the re-prompt are then used to generate the image. Our results illustrate the limited understanding of user prompts by the LISA-13B model. For instance, in the instruction "\textit{Remove the star on the wall}" the key mask area should be around the star, not the entire wall. The absence of a CoT process in LISA-13B results in a weaker understanding and selection of the area to be modified. Moreover, without a re-prompt phase, its inpainting performance is compromised, leading to less accurate content addition.

\begin{figure*}[ht]
\begin{center}
\centerline{\includegraphics[width=0.9\columnwidth]{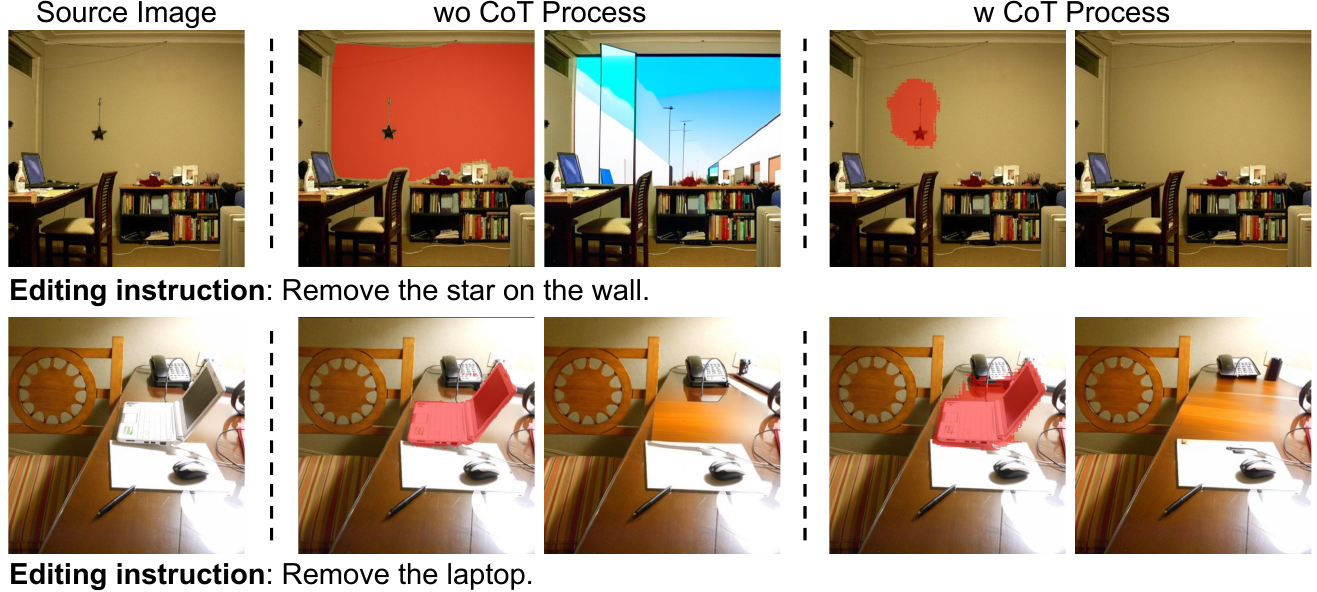}}
\caption{Effectiveness of CoT process.}
\label{fig:abl_cot}
\end{center}
\vskip -0.2in
\end{figure*}

\begin{figure*}[ht]
\vskip 0.2in
\begin{center}
\centerline{\includegraphics[width=1.1\columnwidth]{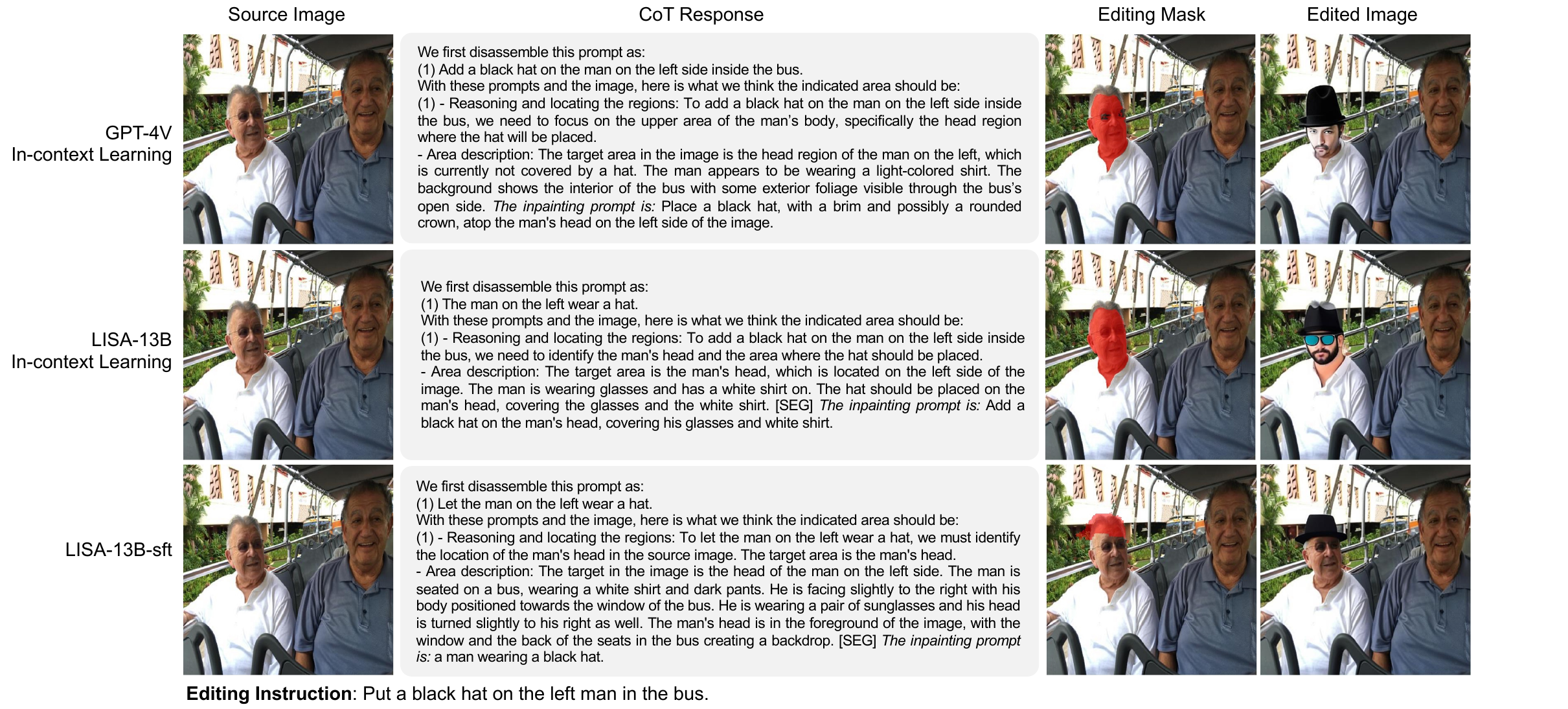}}
\caption{Effectiveness of fine-tuning.}
\label{fig:abl_sft}
\end{center}
\vskip -0.2in
\end{figure*}

\textbf{Effectiveness of fine-tuning.} Three distinct scenarios were considered: (1) the in-context learning ability of the GPT-4V model; (2) the in-context learning ability of the LISA-13B model; and (3) the ability of the LISA-13B model after fine-tuning (denoted as LISA-13B-sft). All models exhibited similar reasoning abilities, effectively describing the mask's specific location and generating effective re-prompt. However, the LISA-13B-sft model excelled in mask creation compared to the GPT-4v and LISA-13B models. As depicted in Fig.\ref{fig:abl_sft}, the LISA-13B-sft model precisely identified the area on the head where a hat should be placed, while the other models outlined the entire head region. This precision is attributed to our joint fine-tuning of the CoT reasoning process and mask. This approach allows the generated mask to contain more relevant information from the editing instruction, pinpointing the exact location to be modified, thus leading to improved mask creation and superior inpainting results. Moreover, the GPT-4V model had a significantly longer inference time than the LISA models, positioning the LISA-13B-sft model as the optimal choice in terms of both performance and efficiency. Further information, including prompts for in-context learning, is in Tab.\ref{tab:ICL}.

\section{Discussion}
\textbf{Data Limitations} The used dataset MagicBrush \cite{NEURIPS2023_64008fa3}, presents certain constraints in both quantity and quality. While this dataset was chosen for its inclusion of editing instructions and associated masks, the masks provided are immature and not refined to the object level. This coarse granularity has resulted in the generation of similarly rough masks with blurred boundaries during model fine-tuning. Additionally, the small size of the dataset poses further restrictions. Future improvements could be achieved by fine-tuning our model with datasets that offer more precise, object-level masks, which would enhance the mask generation capability and allow for more precise image editing.

\textbf{Inpainting Capabilities} The effectiveness of our used inpainting model heavily relies on the quality of the prompt descriptions for the inpainting areas. The precision of these prompts significantly influences the quality of the resulting images, thus we have to carefully design the re-prompting instruction during the CoT process. Moreover, the inherent randomness in the diffusion model's performance can affect the consistency of the image quality. However, a notable advantage of our model is its training-free compatibility with diffusion inpainting processes. This means that should more advanced inpainting models become available, they can be directly integrated into our framework without the need for retraining, potentially enhancing the quality of generated images. This flexibility allows for continual improvement in our model's performance as more sophisticated inpainting technologies are developed.

\section{Conclusion}
In conclusion, our study has addressed two prevalent challenges in traditional text-conditional image models and multimodal large language models (MLLM)-based image generation: the lack of robust understanding of complex instructions and the inability to generate high-fidelity images. We introduced a novel generative framework that bridges MLLM with image generation, leveraging the impressive reasoning capabilities and instance-level segmentation prowess of multimodal LLMs to enhance the controllability of diffusion models. The core of our framework is the prompt reasoning process, employing a fine-tuned LISA model informed by Chain-of-Thought (CoT) process. The second pillar is the LISA localizing process, which generates masks for objects specified within the text. The final phase, the stable diffusion inpainting process, synthesizes images that adhere to the complex prompts while preserving the essence of the source material. Our study's contributions include a comprehensive image generation model, a novel methodology utilizing text generated by MLLM and images generated by LISA to inform the diffusion model, and a new pipeline dataset for advancing image synthesis research. These advancements provide a robust and cost-effective solution for image synthesis from complex textual prompts while ensuring high-fidelity image generation.







{
\small

\bibliographystyle{plain}
\bibliography{neurips_2024}

}

\clearpage

\appendix

\section{Appendix / supplemental material}


\begin{table}[htp]
\centering
\begin{tcolorbox}[colframe=gray, 
                  colback=white!94!black, 
                  arc=4mm, 
                  boxsep=5mm, 
                  boxrule=1pt] 
\textbf{Instruction Decomposition} \\
message=[\{"role": "user", "content": Your task involves deconstructing complex instructions into a sequence of simpler, actionable steps, aiding in better understanding and execution of the task at hand. Each of these simple steps should involve only a single operation: either adding, deleting, or altering an object. Currently, you are provided with complex instructions, denoted as <prompt>. Please parse these instructions into a series of simpler steps.\}]\\

\textbf{Region Localization}\\
message=[\{"role": "user", "content": As an expert in image analysis, your task is to identify the area in the image that corresponds to a given prompt. Your role is to analyze the prompts and determine the specific regions in the image that needs editing. Once you've identified the area, you should clearly indicate them within the image. The prompts provided are as follows: <simple prompts>. \textbackslash n <img>\}]\\

\textbf{Detail description}\\
message=[\{"role": "user", "content": Once the area to be edited has been identified, your task is to provide a detailed area description and a re-prompt that specify the desired changes. The detailed area description should encompass attributes such as the area's relative position within the image, its color, shape, and other notable features. The re-prompt should clearly articulate the intended appearance of the area post-editing. By furnishing this detailed information, you facilitate precise identification of the area in question and clearly define the expected transformation. The results should be in format like this:
- Reasoning and locating the regions:\textbackslash n To add a suitcase in place of the laptop, we need to identify the laptop's location in the source image. The target area to be edited is where the laptop is situated.\textbackslash n- Area description:\textbackslash n The target area in the image is where the laptop is placed, which is on the left side of the desk. The laptop has a black-colored body with white-colored keys on the keyboard. It is positioned near a computer mouse and a phone, with its screen visible and displaying a webpage with red and black elements. The laptop's screen is slightly tilted towards the viewer.\textbackslash n -The inpainting prompt is a suitcase on the leftside of the image.
\textbackslash n <img>\}]
\end{tcolorbox}
\caption{Prompt templates with GPT-4V for instruction decomposition, region localization and detailed description.}
\label{tab:data_generation}
\end{table}

\begin{table}[htp]
\centering
\begin{tcolorbox}[colframe=gray, 
                  colback=white!94!black, 
                  arc=4mm, 
                  boxsep=5mm, 
                  boxrule=1pt] 
\textbf{In-Context Learning Prmopt} \\
message=[\{"role": "user", "content": As an expert in image analysis, your task is to identify the area in the image that corresponds to a given prompt. Your role is to analyze the prompts and determine the specific regions in the image that needs editing. Once you've identified the area, you should clearly indicate them within the image. The prompts provided are as follows: Add a black hat on the man on the left side inside the bus. \\
Once the area to be edited has been identified, your task is to provide a detailed area description and a re-prompt that specify the desired changes. The detailed area description should encompass attributes such as the area's relative position within the image, its color, shape, and other notable features. The re-prompt should clearly articulate the intended appearance of the area post-editing. By furnishing this detailed information, you facilitate precise identification of the area in question and clearly define the expected transformation. \\
The results should be in format like this:
- Reasoning and locating the regions:\textbackslash n To add a suitcase in place of the laptop, we need to identify the laptop's location in the source image. The target area to be edited is where the laptop is situated.\textbackslash n \textbackslash n- Area description:\textbackslash n The target area in the image is where the laptop is placed, which is on the left side of the desk. The laptop has a black-colored body with white-colored keys on the keyboard. It is positioned near a computer mouse and a phone, with its screen visible and displaying a webpage with red and black elements. The laptop's screen is slightly tilted towards the viewer.\textbackslash n \textbackslash n-Reprompt:\textbackslash n a suitcase on the leftside of the image.\}]
\end{tcolorbox}
\caption{Prompt template for GPT-4V in-context learning and LISA-13B in-context learning .}
\label{tab:ICL}
\end{table}

\end{document}